\pdfoutput=1

\documentclass[11pt,a4paper]{article}
\usepackage{acl}
\usepackage{times}
\usepackage{booktabs}
\usepackage[utf8]{inputenc}
\usepackage{graphicx}
\usepackage{hyperref}
\usepackage{breqn}
\usepackage{booktabs}
\usepackage{todonotes}
\usepackage{adjustbox}
\usepackage{array}
\newcolumntype{H}{>{\setbox0=\hbox\bgroup}c<{\egroup}@{}}
\usepackage{float}

\usepackage{microtype}

\usepackage{enumitem}
\setlist{nosep}

\title{Do Children Texts Hold The Key To Commonsense Knowledge?}
\author{Julien Romero \\
  Télécom SudParis \\
  \texttt{jromero@telecom-sudparis.eu} \\\And
  Simon Razniewski \\
  Max Planck Institute for Informatics \\
  \texttt{srazniew@mpi-inf.mpg.de} \\}

\begin{document}

\maketitle

\begin{abstract}
Compiling comprehensive repositories of commonsense knowledge is a long-standing problem in AI. Many concerns revolve around the issue of \textit{reporting bias}, i.e., that frequency in text sources is not a good proxy for relevance or truth. This paper explores whether children's texts hold the key to commonsense knowledge compilation, based on the hypothesis that such content makes fewer assumptions on the reader's knowledge, and therefore spells out commonsense more explicitly. An analysis with several corpora shows that children's texts indeed contain much more, and more typical commonsense assertions. Moreover, experiments show that this advantage can be leveraged in popular language-model-based commonsense knowledge extraction settings, where task-unspecific fine-tuning on small amounts of children texts (childBERT) already yields significant improvements. This provides a refreshing perspective different from the common trend of deriving progress from ever larger models and corpora.
\end{abstract}

\section{Introduction}

Compiling commonsense knowledge (CSK) is a long-standing problem in AI~\cite{lenat1995cyc}. Automated text-extraction-based approaches to CSK compilation, like Knext \cite{DBLP:conf/aaai/GordonDS10}, TupleKB \cite{tuplekb}, Quasimodo \cite{quasimodo}, COMET \cite{hwang2020comet} or Ascent \cite{ascent} typically struggle with \textit{reporting bias} \cite{reportingbias,mehrabi2021lawyers}, in particular an under-reporting of basic commonsense assertions. This is a crux of commonsense: If knowledge is assumed to be commonplace, such as that \emph{rain is wet} or \emph{cars have wheels}, there is little need to utter it explicitly. In contrast, statements that contradict commonsense are more frequently reported, leading to inappropriate images of the real world, e.g., that fires are more often cold than hot (e.g., 238 vs.\ 173 literal occurrences in the English Wikipedia).

Children's material may partially counter this bias: As children's knowledge is still growing, seemingly obvious assertions may still be frequently expressed explicitly in such material. Note that this is not a binary question of whether some knowledge is expressed or not, but more a ranking problem: Prominent CSK repositories often do not struggle to recall relevant statements (e.g., Ascent \cite{ascent} contains 2800 assertions for ``elephant''), but struggle to rank them properly. This is especially true for language-model based approaches of CSK compilation \cite{hwang2020comet,west2021symbolic}, which by design can assign every token in the vocabulary a probability, but should do so in sensible order.

This paper investigates (i) whether children's texts are a promising source for CSK and (ii) whether small corpora can still boost knowledge extraction from large language models. Specifically, we analyze the density and typicality of CSK assertions in children's text corpora and show how fine-tuning existing language models on them can improve CSK compilation. Data and models, including a childBERT variant, can be found at \url{https://www.mpi-inf.mpg.de/children-texts-for-commonsense}.
\section{Background}

Prominent manual efforts towards CSK compilation include ConceptNet~\cite{speer2017conceptnet}, Atomic \cite{atomic}, and the integrated CSKG \cite{ilievski2021cskg}. 
Prominent text extraction projects are Knext~\cite{DBLP:conf/aaai/GordonDS10}, TupleKB~\cite{tuplekb}, Quasimodo~\cite{quasimodo} and Ascent \cite{ascentpp}. Each carefully selects extraction corpora, like Wikipedia texts, user query logs, or targeted web search, to minimize extraction noise and maximize salience. Nonetheless, all struggle with extracting very basic CSK that is generally deemed too obvious to state explicitly.
The utilized corpora are also small compared to what is typically used in language model pre-training. Therefore, pre-trained language models (PTLMs) have been employed directly for CSK extraction in a setting called prompting/probing (cf.\ the LAMA benchmark) \cite{petroni2019language}, where the BERT LM showed promising results in predicting ConceptNet assertions. They can also be employed with supervision, like in the COMET and the Atomic$^{10x}$ system \cite{hwang2020comet,west2021symbolic}. However, both PTLM-paradigms are grounded in frequencies observed in the original text corpora used for LM training, which are again subject to reporting bias.

\section{Children Text Corpora}

For understanding the nature of different text corpora, we rely on the Flesch Reading-ease score (FRE)~\cite{flesch1979write} that is based on the number of syllables, words, and sentences.

It generally ranges between 0 and 100, with 0-30 being considered difficult to read, 60-70 assumed standard, and above 80 easy.

We investigate three children text corpora:
\begin{enumerate}
    \item \textbf{Children Book Test (CBT)} The CBT dataset~\cite{hill2015goldilocks} contains 108 children books such as \textit{Alice's Adventures in Wonderland} extracted from the \href{https://www.gutenberg.org}{Gutenberg Project}. It targets children around 12-14 years old and is about 30 MB in total.
    \item \textbf{C4-easy} C4~\cite{2019t5} is a cleaned version of Common Crawl's web crawl corpus that was used to train the T5 language model. It is approximately 305 GB in size. We derive \textit{C4-easy} by restricting the corpus to documents with an FRE greater than 80, retaining 40.827.011 documents, which are 11\% of C4.
    \item \textbf{InfantBooks} We newly introduce the InfantBooks dataset, composed of 496 books targeted at kids from 1-6 years. It is based on Ebooks from websites like \href{https://freeInfantBooks.org/}{freekidsbooks.org}, \href{https://monkeypen.com/}{monkeypen.com} and \href{https://www.kidsworldfun.com}{kidsworldfun.com}, which we collected, transcribed, and cleaned. The final dataset consists of 496 books with 2 MB of text.\footnote{The dataset is available at \url{https://www.mpi-inf.mpg.de/children-texts-for-commonsense}.}
\end{enumerate}
As a baseline, and to rule out that observed improvements stem only from general training on more data, we also compare with employing the whole C4 corpus. 
%
%
In Table~\ref{tab:dataset-statistics}, we compare the corpora according to average document length, vocabulary size, and readability. In Table~\ref{tab:dataset-statistics2}, we make the same comparison with the number of distinct words, the number of frequent words (with a relative frequency greater than 0.01\%), and the cumulative frequency of the top 1000 words.

\begin{table}[ht]
\small
\centering
\begin{adjustbox}{width=\columnwidth,center}
\begin{tabular}{|l|Hl|l|l|}
\hline
\textbf{Corpus} & \textbf{Size} & \textbf{Avg.\ doc.\ len.} & \textbf{Vocab.\ size} & \textbf{Readability (FRE)} \\
\hline
\textit{C4} & \textit{20 sent. avg.} & \textit{411 words} & \textit{151k} & \textit{60 (Standard)} \\

CBT & 2721 sent. avg. (108 books) \ & 57k words & 63k & 62 (Standard) \\
C4-easy & 22 sentences in average & 317 words & 106k & 86 (Easy) \\
InfantBooks & avg. 55 sent. avg. (496 books)\ & 659 words & 18k & 91 (Very Easy) \\
\hline
\end{tabular}
\end{adjustbox}
\caption{Text corpora considered for pretraining/finetuning, sorted by FRE.}
\label{tab:dataset-statistics}
\end{table}

\begin{table}[ht]
\small
\centering
\begin{adjustbox}{width=\columnwidth,center}
\begin{tabular}{|l|l|l|l|}
\hline
\textbf{Corpus} & \textbf{Dist. Words} & \textbf{freq. words} & \textbf{Cumul. freq. top 1k} \\
\hline
\textit{C4} & 8M \ & 994 & 68\% \\

CBT & 5M & 874 & 82\% \\
C4-easy & 8M & 908 & 75\% \\
InfantBooks & 5M & 1031 & 82\% \\
\hline
\end{tabular}
\end{adjustbox}
\caption{Text corpora statistics.}
\label{tab:dataset-statistics2}
\end{table}

\section{Analysis}

\paragraph{CSK Density.}

Although CBT and InfantBooks are too small for comprehensive text extraction, it is informative to see how dense CSK assertions are stated in them, i.e., the relative frequencies of CSK assertions per text. 

We used the CSLB~\cite{cslb} dataset, a large crowdsourced set of basic CSK assertions, like \textit{alligator: is scary / is long / is green}. We focused on the top 4,245 properties for 638 subjects stated at least five times. For each corpus, we computed the relative frequencies with which these statements appear (w/ lemmatization).


Table~\ref{tab:direct_object_property} shows the results. As one can see, InfantBooks has the highest relative density of CSK assertions, 3x as many as C4 per sentence, 5x more per word.

\begin{figure}
    \centering
    \includegraphics[width=0.5\textwidth]{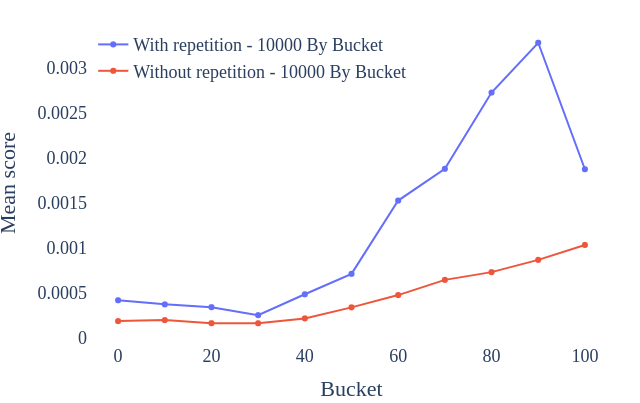}
    \caption{Relative CSK assertion frequency (per word) by C4 bucket of different readability (FRE).}
    \label{fig:bucket_words}
\end{figure}

To further explore the relation of text simplicity and CSK density, we grouped C4 documents into buckets based on their FRE.
For a sample of 10k documents per bucket, Figure~\ref{fig:bucket_words} reports the per-word frequencies of CSK assertions, considering all spotted CSK assertions (blue) or only distinct ones (red). The results are shown in Figure~\ref{fig:bucket_words}. As one can see, CSK density increases significantly with easier readability, and only the most simple documents suffer from a lack of diversity (decrease in blue line).

\begin{table}[ht]
     \centering
     \small
     \begin{tabular}{|c|c|c|c|c|}
        \hline
            \textbf{Dataset} & \textbf{Mean freq./sent.} & \textbf{Mean freq./word} \\
            \hline
            C4 (sample) & 9.99e$^{-3}$ & 4.86e$^{-4}$ \\
            CBT & 1.03e$^{-2}$ & 4.94e$^{-4}$ \\
            C4-easy (sample) & 1.10e$^{-2}$ & 7.65e$^{-4}$ \\
            InfantBooks & \textbf{3.07e$^{-2}$} & \textbf{2.56e$^{-3}$}\\
            \hline
        \end{tabular}
     \caption{CSK density in different corpora.}
     \label{tab:direct_object_property}
 \end{table}



\paragraph{CSK Typicality.}
Human associations and CSK resources contain a mix of salient but atypical assertions (\textit{lion attacks human}), and typical but basic assertions (\textit{lion drinks water}) \cite{chalier2020joint}. To evaluate whether children's texts can help extract more typical assertions, we use PTLM prompting. Prompting is a standard procedure for extracting knowledge from masked-LMs~\cite{liu2021pre}. Given a relation of interest, e.g., ``LocatedAt'', one uses a generic textual pattern like ``$<$entity$>$ can be found at [MASK]'', to obtain suggestions from an LM.

We used the BERT-large model\footnote{\href{https://huggingface.co/bert-large-uncased}{https://huggingface.co/bert-large-uncased}}, and fine-tuned it on each of the four corpora, up to a maximum of 48 hours (using one NVIDIA Quadro RTX 8000). We used a learning rate of $1e^{-5}$.

We then used 1,600 CSK triples from Quasimodo \cite{quasimodo}, human-rated as \textit{very typical} (score [1,2[), or \textit{typical} (score [2,3[), or \textit{plausible} (score [3,4[) as targets. For each of these triples, we masked the objects, and used the fine-tuned BERT model to predict token likelihoods for the mask.

\begin{table}[ht]
     \centering
     \small
    \begin{tabular}{|c|c|c|c|}
    \hline
         \textbf{BERT-finetuning} & \textbf{Very typical} & \textbf{Typical} & \textbf{Plausible} \\
         \hline
         None & 0.361 & 0.200 & 0.153 \\
         InfantBooks & \textbf{0.364} & 0.198 & 0.152\\
         CBT & \textbf{0.364} & \textbf{0.205} & 0.148\\
         C4-easy & 0.326 & 0.204 & \textbf{0.156}\\
         C4 & 0.312 & 0.195 & 0.155\\
         \hline
    \end{tabular}
    \caption{MRR of typical statements.}
    \label{tab:typicality}
\end{table}

In Table \ref{tab:typicality} we report the mean reciprocal rank (MRR) for each group. Pre-training on children corpora has a slight edge over the vanilla BERT model and a significant edge over pre-training on the general C4 corpus.

\section{CSK Generation}

We next evaluate whether pre-training on children texts helps in two PTLM-based methods for CSK generation: (i) via prompting from a LM pretrained on those corpora (LAMA method) \cite{petroni2019language}, and (ii) via supervised learning, based on a LM fine-tuned on those corpora (COMET method) \cite{hwang2020comet}.

\paragraph{Unsupervised Generation via PTLM Prompting.}
\label{sec:lama}


%
The LAMA probe is based on prompting-based (CSK) generation. Originally, it used ConceptNet~\cite{speer2017conceptnet} together with the Open Mind Common Sense~\cite{singh2002open} (OMCS), which was used to construct ConceptNet and contains 20k items. ConceptNet contains triplets, and each triplet is associated with a sentence in OMCS. When a triplet has an object composed of one token, then the object is masked in the original OMCS sentence. The problem with this approach is that OMCS sentences are unnatural template-based phrases, not suited for LM completion. For example, in the LAMA probe, we find that ``One of the things you do when you are alive is [MASK].'' (think), ``Something that might happen while analyzing something is [MASK].'' (education) or ``Similarity between a trash container and a clock: both can be found in a [MASK].'' (school).

We therefore decided to add new commonsense-related datasets, to better evaluate how much commonsense knowledge a language model contains. First, we used CSLB~\cite{cslb} that contains 20k basic properties about given subjects. The sentences are generally simpler than in OMCS. Second, we exploit 2.4k human-generated CSK sentences produced as evaluation data in the Quasimodo project~\cite{quasimodo,romero2020inside}. Finally, we exploited sentence sources of top statements web-extracted for the same KB, based on QA forums and search engine query logs. We here either masked the object or the predicate for all the sentences. 
Table~\ref{tab:lama_probe_samples} shows a sample of probes for each dataset. We also make public (link above) a sample of generations.

\begin{table}[ht]
     \centering
     \small
     \begin{tabular}{|c|c|}
        \hline
        \textbf{Dataset} & \textbf{Sample Probes} \\
        \hline
        ConceptNet & \begin{tabular}{@{}c@{}}Something that might happen while \\ fiddling is [MASK] a string. \textbf{breaking} \\ Geriatrics is a type of [MASK]. \textbf{nursing} \\ A person doesn't want to \\ be [MASK]. \textbf{crippled} 
        \end{tabular} \\
        \hline
        CSLB & \begin{tabular}{@{}c@{}}Dolphin is an [MASK]. \textbf{animal} \\ Mug hold [MASK]. \textbf{tea} \\ Cat chase [MASK]. \textbf{strings} \end{tabular}\\
        \hline
        Q'modo-eval & \begin{tabular}{@{}c@{}} Salmons are [MASK]. \textbf{fish} \\ Ducks [MASK] in water. \textbf{live} \\ Barbers cut our [MASK]. \textbf{hair} \end{tabular}\\
        \hline
        Quasimodo &  \begin{tabular}{@{}c@{}}Pencils are made of [MASK]. \textbf{graphite} \\ Bumblebees collect [MASK]. \textbf{pollen} \\ Rockets need fuel in [MASK]. \textbf{space}\end{tabular}\\
        \hline
    \end{tabular}
     \caption{Sample prompts for LAMA-style evaluation.}
     \label{tab:lama_probe_samples}
 \end{table}


\begin{table*}[ht]
     \centering
     \begin{adjustbox}{width=2\columnwidth,center}
     \begin{tabular}{|c|c|c|c|c|c|c|c|c|c|c|c|c|}
        \hline
        \textbf{Corpus} & \multicolumn{3}{|c|}{\textbf{ConceptNet}} & \multicolumn{3}{|c|}{\textbf{CSLB}} & \multicolumn{3}{|c|}{\textbf{Quasimodo-eval} } & \multicolumn{3}{|c|}{\textbf{Quasimodo}} \\
        \hline
        & \textbf{MRR} & \textbf{Hits@1} & \textbf{Hits@10} & \textbf{MRR} & \textbf{Hits@1} & \textbf{Hits@10} & \textbf{MRR} & \textbf{Hits@1} & \textbf{Hits@10} & \textbf{MRR} & \textbf{Hits@1} & \textbf{Hits@10}\\
        \hline
        None & \textbf{0.270$^*$} & \textbf{0.188$^*$} & \textbf{0.429} & 0.149 & 0.0911 & 0.259 & 0.433 & 0.327 & 0.633 & 0.212 & 0.138 & 0.357 \\
        InfantBooks & 0.260 & 0.177 & 0.421 & \textbf{0.174$^*$} & \textbf{0.106$^\dagger$} & \textbf{0.317$^*$} & \textbf{0.471$^\dagger$} & \textbf{0.374$^\dagger$} & \textbf{0.652} & \textbf{0.231$^*$} & \textbf{0.146$^\dagger$} & \textbf{0.404$^*$} \\
        CBT & 0.259 & 0.177 & 0.414 & 0.163 & 0.100 & 0.285 & 0.457 & 0.355 & 0.642 & 0.223 & 0.142 & 0.387 \\
        C4-easy & 0.243 & 0.165 & 0.395 & 0.150 & 0.0877 & 0.274 & 0.390 & 0.288 & 0.586 & 0.200 & 0.123 & 0.356\\
        C4 & 0.229 & 0.155 & 0.372 & 0.137 & 0.0777 & 0.251 & 0.457 & 0.356 & 0.637 & 0.182 & 0.111 & 0.325 \\
        \hline
    \end{tabular}
\end{adjustbox}

     \caption{LAMA-style evaluation of CSK generation. ($^*$: significantly better than all others with p-value$<$0.05, $^\dagger$: significantly better than having no corpus with p-value$<$0.05)}
     \label{tab:lama_probe}
 \end{table*}

\begin{table*}[ht]
    \centering
    \small
    \begin{adjustbox}{width=2\columnwidth,center}
    \begin{tabular}{|p{.265\linewidth}|p{.34\linewidth}|p{.3\linewidth}|}
        \hline
        \textbf{Masked Sentence} & \textbf{BERT-large} & \textbf{BERT-large finetuned on InfantBooks}  \\
        \hline
        Bears [MASK]. & Club, \#\#kin, \#\#eed,\#\#k, !, vs, \#\#ki, \#\#ville, Town, \#\#t & \textbf{bite}, \textbf{eat}, \textbf{walk}, sing, do, \#\#t, dance, say, \textbf{sleep}, kiss \\
        \hline
        Surgeons treat [MASK]. & patients, them, children, wounds, animals, prisoners, \textbf{victims}, \textbf{people}, him, dogs & children, \textbf{pain}, wounds, \textbf{cold}, dogs, \textbf{burns}, birds, patients, \textbf{babies}, cats\\
        \hline
        Chefs know a lot about [MASK]. & cooking, food, it, that, them, wine, this, you, fish, me & cooking, food, fish, wine, \textbf{baking}, \textbf{meat}, \textbf{recipes}, it, \textbf{dishes}, them \\
        \hline
        Researchers study [MASK]. & it, them, this, children, women, him, animals, evolution, her, birds & animals, \textbf{art}, birds, \textbf{chemistry}, \textbf{insects}, it, \textbf{music}, \textbf{snakes}, \textbf{science}, them \\
        \hline
    \end{tabular}
    \end{adjustbox}
    \caption{Sample Top Predictions (\#\# means the letters are concatenated to the previous word). In bold good predictions that did not appear in the other method.}
    \label{tab:sample_lama}
\end{table*}

We used the BERT-large model pretrained as is, or pre-trained on any of the children's text corpora, as in the previous section, to generate predictions for each of the masked probes. The resulting performances are reported in Table~\ref{tab:lama_probe}. We report MRR and Hits@k for the PTLM's predictions in each case. While there are no improvements on the idiosyncratic ConceptNet data, pre-training on InfantBooks performs consistently at the top in the three other settings. CBT also shows consistent gains, while the broader C4-easy helps little. Examples are shown in Table~\ref{tab:sample_lama}.

\paragraph{Supervised PTLM-based Generation.}

Beyond prompting, an established technique to obtain CSK assertions is supervised learning based on LMs. We adapted the COMET-ATOMIC-2020 system~\cite{hwang2020comet}, which in turn is based on the GPT-2 LM. Like the previous prompt-based extraction, the core component is a pre-trained LM, which we can fine-tune on different text corpora. Compared to the previous method, however, a relation-specific step of supervised learning is added, in which the model is trained to generate objects for given subject-relation pairs.

We used ConceptNet for training and testing\footnote{Observe that in the previous unsupervised setting, the other datasets were needed because the \textit{source sentences} underlying ConceptNet are unnatural. Its structured statements themselves are of good quality.} and report precision and recall@k. The training and testing sets contain 
186k and 23k triples.


\begin{table}[ht]
    \small
    \centering
    \begin{tabular}{|c|c|c|c|c|c}
    \hline
        \textbf{Corpus} & \textbf{P@5} & \textbf{P@10} & \textbf{R@5} & \textbf{R@10} \\
        \hline
        None & 3.76\% & 2.51\% & 15.4\% & 20.0\% \\
        C4-easy & 3.78\% & 2.48\% & 15.5\% & 19.9\% \\
        C4 & 3.73\% & 2.47\% & 15.3\% & 19.7\% \\
        CBT & 3.74\% & 2.48\% & 15.4\% & 19.9\% \\
        InfantBooks\!\! & \textbf{3.81\%} & \textbf{2.52\%} & \textbf{15.6\%} & \textbf{20.1\%} \\
        \hline
    \end{tabular}
    \caption{Supervised COMET-style generation of CSK.}
    \label{tab:comet-results}
\end{table}

The results are shown in Table~\ref{tab:comet-results}. Due to the additional supervision on KB statements, differences are much smaller than in the previous setting. Nonetheless, we observe a consistent edge for fine-tuning on InfantBooks. To confirm this, we employed an additional human evaluation. For each subject-predicate pair in the test dataset, we computed the top-1 prediction for each model. For 300 examples where the predictions differed, we asked human annotators for their pairwise preference in terms of correctness. The results are shown in Table~\ref{tab:manual_annotation_comet}. InfantBooks again outperforms the vanilla BERT model and the more general C4-easy. Table~\ref{tab:sample_comet} shows examples of generations.

We also performed an absolute evaluation of 600 pairs: For each pair, we asked a human evaluator to score the similarity between 1 (the worst) and 5 (the best). We obtained an average typicality of 2.57 for COMET-InfantBooks and 2.53 for COMET-C4. This is consistent with the results of the relative analysis - the gains in the KB-supervised settings are small, but consistently observable both in absolute and relative terms.

\begin{table}[!ht]
    \centering
    \small
    \begin{tabular}{|c|c|c|}
    \hline
    \textbf{InfantBooks Best} & \textbf{BERT Best} & \textbf{Same} \\
        \hline
        \textbf{13\%} & 10\% & 77\% \\
        \hline
        \textbf{InfantBooks Best} & \textbf{C4 Easy Best} & \textbf{Same} \\
        \hline
        \textbf{22\%} & 15\% & 63\% \\

    \hline
    \end{tabular}
    \caption{Human preference of CSK generations.} 
    \label{tab:manual_annotation_comet}
\end{table}

\begin{table}[h]
    \centering
    \small
    \begin{adjustbox}{width=\columnwidth,center}
    \begin{tabular}{|p{.5\linewidth}|p{.25\linewidth}|p{.3\linewidth}|}
        \hline
        \textbf{Input (SP)} & \textbf{BERT-large} & \textbf{BERT-large finetuned on InfantBooks}  \\
        \hline
        bill, CapableOf & pay bill & amount nothing \\
        standing up, HasSubevent & getting dizzy & falling down \\
        hair, AtLocation & cabinet & hair salon \\
        creating idea, Causes & solution & new idea \\
        chew food, MotivatedByGoal & tastes good & eat \\
        \hline
    \end{tabular}
    \end{adjustbox}
    \caption{Sample COMET Predictions}
    \label{tab:sample_comet}
\end{table}

\section{Conclusion}

Our results positively confirm both starting questions: Dedicated children corpora contain more typical CSK assertions, and even small quality text corpora can boost the performance of large LMs for CSK extraction, especially concerning basic CSK assertions.

While the overall gains are still modest, these results affirm the role of data selection in commonsense knowledge compilation. Along with the exploitation of even stronger pre-trained language models, the quest for relevant text corpora thus continues.

\section{Limitations}

The following are notable limitations of our study, as well as of our approach.

\paragraph{Size of InfantBooks corpus} A clear limitation of our newly presented InfantBooks corpus is its small size. Even if we were to ramp up the number of books (496 currently), text length in such books is inherently small, so this corpus will size-wise never compete with popular general web corpora. We alleviated this problem by a solution involving fine-tuning LMs trained on general corpora, but it remains that the presented corpus is not suitable for direct extraction tasks.

\paragraph{Flesch reading-ease vs.\ children content} To further alleviate the size problem of the InfantBooks corpus, we also introduced the C4-easy corpus, based on filtering via the Flesch reading-ease score (FRE). However, good readability is more of a necessary than a sufficient condition for inferring that content is intended for children, in other words, if there were not the results on InfantBooks, the positive results on C4-easy in isolation only proved a related hypothesis (``easier texts are the key to commonsense''), not the original one (``children texts are the key'').

\paragraph{Dataset long-term availability} InfantBooks contains copyrighted content. While we have checked national legislation, and believe that sharing the material for research purposes is permitted, should legal complaints occur, they would pose a risk to our ability to share this material long-term. 

\subsection*{Acknowledgment}
We thank Gerhard Weikum and Tuan-Phong Nguyen for discussions in earlier stages of this work, and the anonymous reviewers for their helpful comments.

\bibliography{refs}
\bibliographystyle{acl_natbib}








\end{document}